\documentclass{article}

\usepackage[final]{nips_2018}

\usepackage[utf8]{inputenc} 
\usepackage[T1]{fontenc}    
\usepackage{hyperref}       
\usepackage{url}            
\usepackage{booktabs}       
\usepackage{amsfonts}       
\usepackage{nicefrac}       
\usepackage{microtype}      
\usepackage{graphicx}
\usepackage{caption}

\title{Training on Art Composition Attributes to Influence CycleGAN Art Generation}

\author{
  Holly Grimm \\
  Independent Software Developer and Artist\\
  \texttt{holly.grimm@gmail.com} \\
}

\begin{document}

\maketitle

\begin{abstract}
 I consider how to influence CycleGAN, image-to-image translation, by using additional constraints from a neural network trained on art composition attributes. I show how I trained the the Art Composition Attributes Network (ACAN) by incorporating domain knowledge based on the rules of art evaluation and the result of applying each art composition attribute to apple2orange image translation.
\end{abstract}

\section{Introduction}

The standard adversarial and cyclical losses of a CycleGAN [1] were augmented with additional loss terms from a convolutional neural network trained with art composition attributes. During training of the CycleGAN, the user specifies values for each of the art composition attributes. For instance, if a target contrast value of 10 is specified, the generator should output images with more contrast than if the target contrast value is 1.

\subsection{Art Composition Attributes}

\begin{figure}[h]
	\centering
	\begin{minipage}{.24\textwidth}
		\centering
		\includegraphics[width=.95\linewidth]{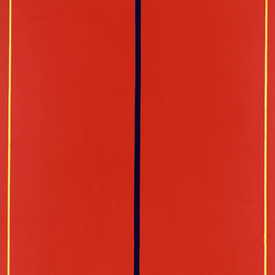}
		\captionof{figure}{low texture}
		\label{fig:lowtextureart}
	\end{minipage}
	\begin{minipage}{.24\textwidth}
		\centering
		\includegraphics[width=.95\linewidth]{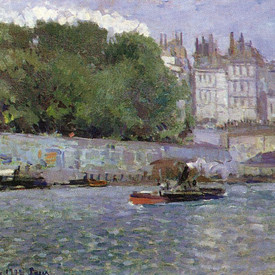}
		\captionof{figure}{high texture}
		\label{fig:hightextureart}
	\end{minipage}
		\begin{minipage}{.24\textwidth}
		\centering
		\includegraphics[width=.95\linewidth]{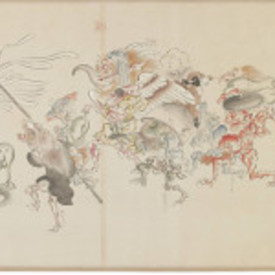}
		\captionof{figure}{low contrast}
		\label{fig:lowcontrastart}
	\end{minipage}
	\begin{minipage}{.24\textwidth}
		\centering
		\includegraphics[width=.95\linewidth]{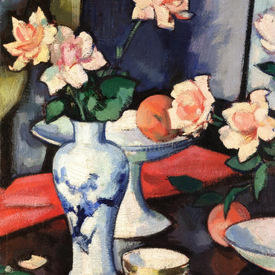}
		\captionof{figure}{high contrast}
		\label{fig:highcontrastart}
	\end{minipage}
\end{figure}

Eight art composition attributes were selected:
variety of texture, variety of shape, variety of size, variety of color, contrast, repetition, primary color, and color harmony. Five hundred images from the WikiArt dataset [2] were labeled with these attributes. Figures \ref{fig:lowtextureart}, \ref{fig:hightextureart}, \ref{fig:lowcontrastart} and \ref{fig:highcontrastart} are examples of low and high values for variety of texture and contrast.

\section{ACAN}
\label{acan}

Training consisted of fine-tuning a ResNet50 [3] pretrained on the ImageNet dataset. ResNet50 is a fifty-layer deep residual network with 16 residual blocks. Global Average Pooling (GAP) is applied to the ReLU output from each of the sixteen ResNet block activations, called rectified convolution maps [4]. The sixteen GAP outputs were concatenated and L2 normalization was applied to create a merge layer. From the merge layer, there are eight outputs, one for each of the attributes.

\section{CycleGAN and ACAN}
\label{cyclegan_and_acan}

In addition to the standard CycleGAN losses (Adversarial, Cycle-Consistency, and Identity) the ACAN losses are a series of eight losses generated when the translated image is passed through the ACAN with eight target values. The difference between these target values and the values output by the network are the attribute losses.

\section{Results}
\label{results}

Below is a sampling of the results of running the CycleGAN training with ACAN on the apple2orange dataset. Even with a small training set size of 500 images, the ACAN is able to learn and generate apples with the eight art compositional attributes.

\begin{figure}[h]
	\centering
	\begin{minipage}{.245\textwidth}
		\centering
		\includegraphics[width=.95\linewidth]{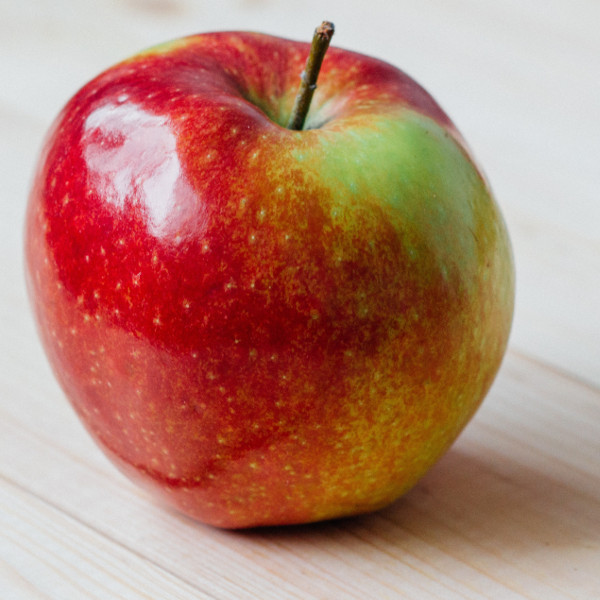}
		\captionof{figure}{original apple image}
		\label{fig:original}
	\end{minipage}	
	\begin{minipage}{.245\textwidth}
		\centering
		\includegraphics[width=.95\linewidth]{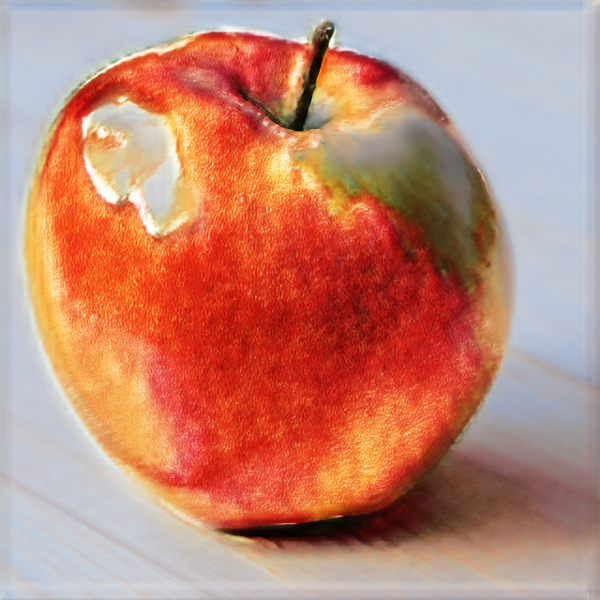}
		\captionof{figure}{color harmony analogous}
		\label{fig:analogous}
	\end{minipage}
	\begin{minipage}{.245\textwidth}
		\centering
		\includegraphics[width=.95\linewidth]{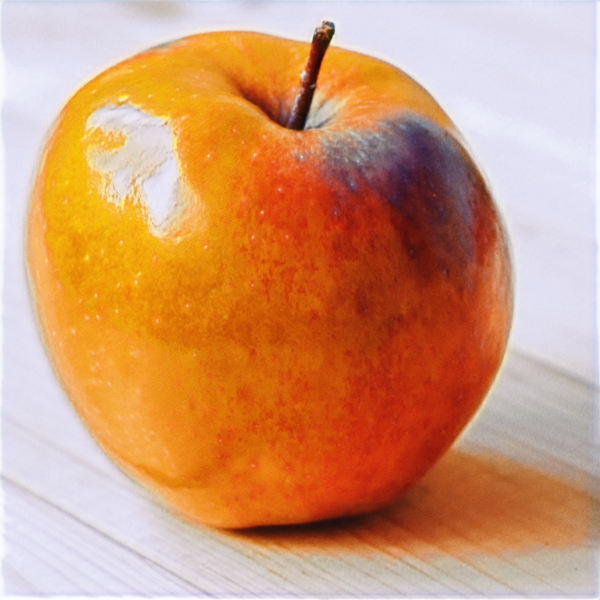}
		\captionof{figure}{color harmony complementary}
		\label{fig:complementary}
	\end{minipage}
		\begin{minipage}{.245\textwidth}
		\centering
		\includegraphics[width=.95\linewidth]{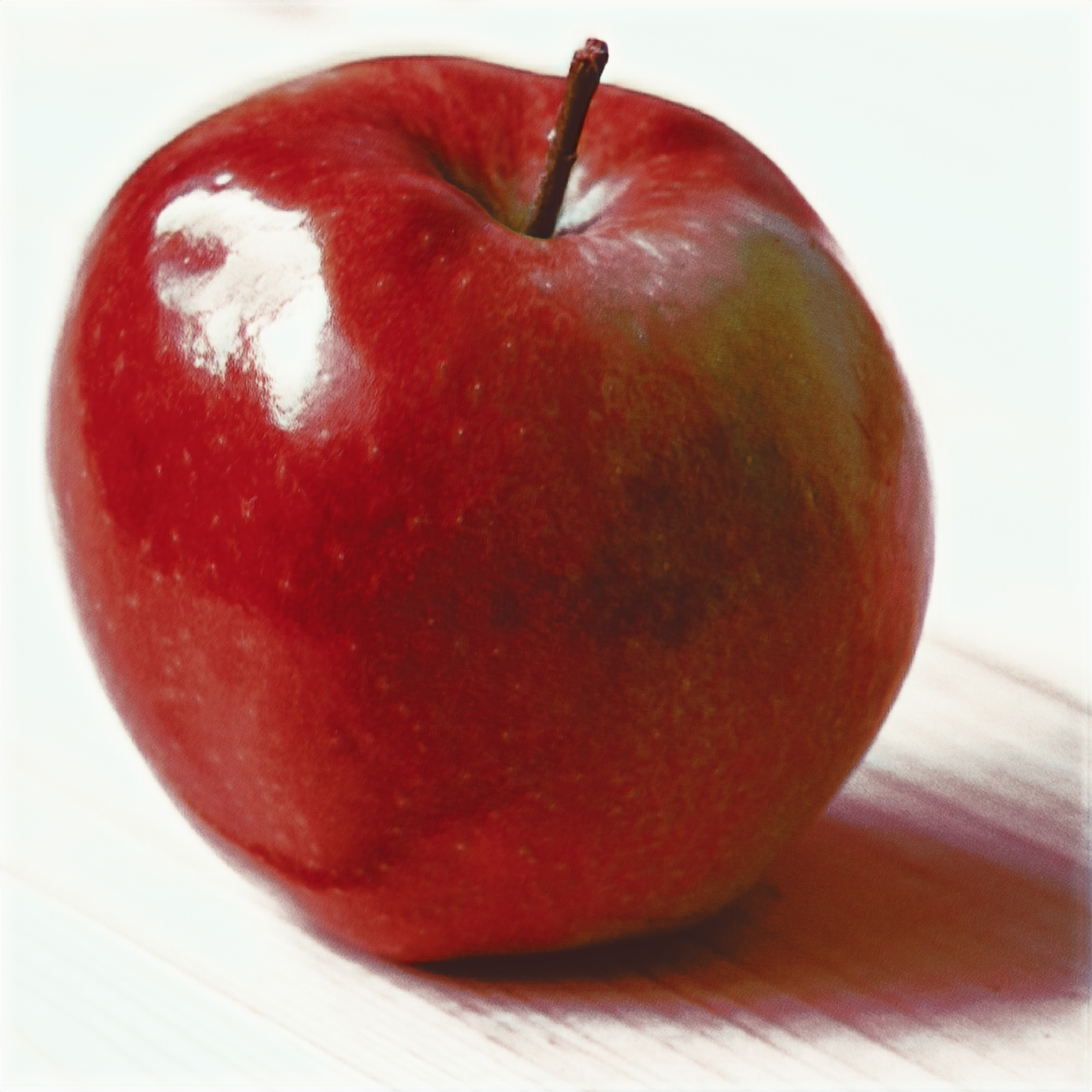}
		\captionof{figure}{color harmony monochromatic}
		\label{fig:monochromatic}
	\end{minipage}
	\begin{minipage}{.245\textwidth}
		\centering
		\includegraphics[width=.95\linewidth]{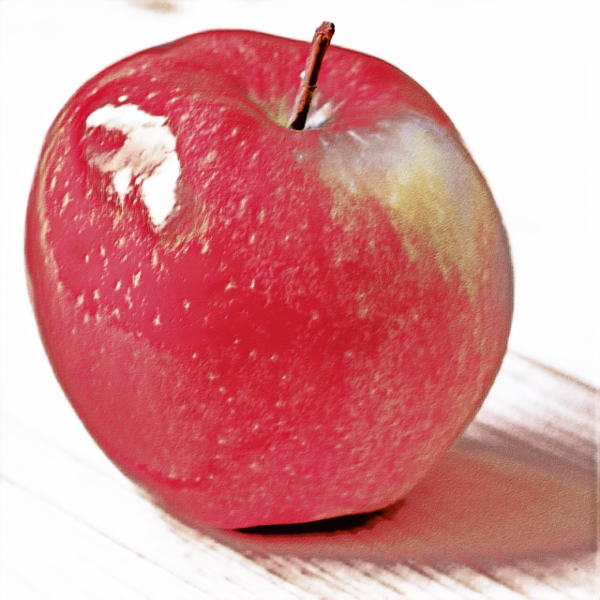}
		\captionof{figure}{low contrast}
		\label{fig:lowcontrast}
	\end{minipage}
	\begin{minipage}{.245\textwidth}
		\centering
		\includegraphics[width=.95\linewidth]{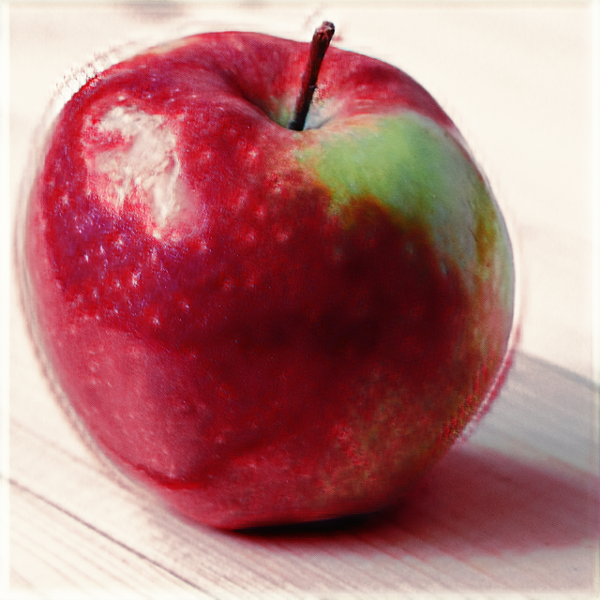}
		\captionof{figure}{high contrast}
		\label{fig:highcontrast}
	\end{minipage}
		\begin{minipage}{.245\textwidth}
		\centering
		\includegraphics[width=.95\linewidth]{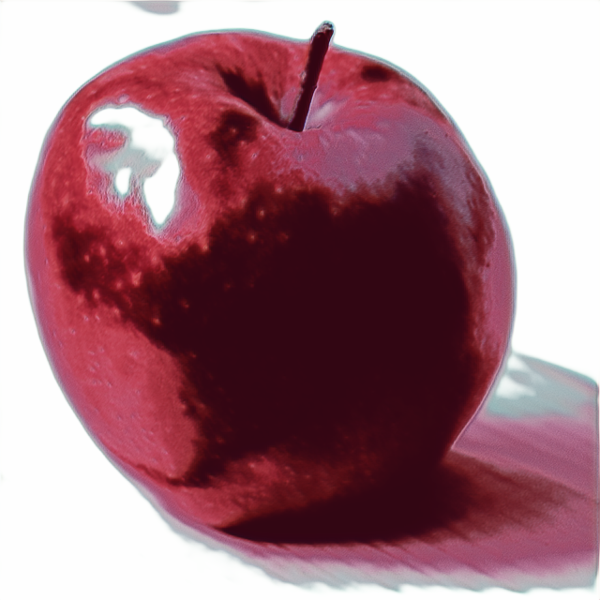}
		\captionof{figure}{low color}
		\label{fig:lowcolor}
	\end{minipage}
	\begin{minipage}{.245\textwidth}
		\centering
		\includegraphics[width=.95\linewidth]{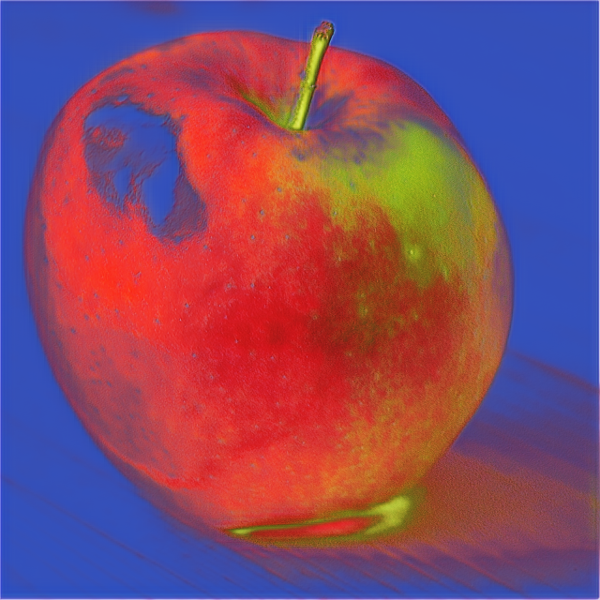}
		\captionof{figure}{high color}
		\label{fig:highcolor}
	\end{minipage}
	\begin{minipage}{.245\textwidth}
		\centering
		\includegraphics[width=.95\linewidth]{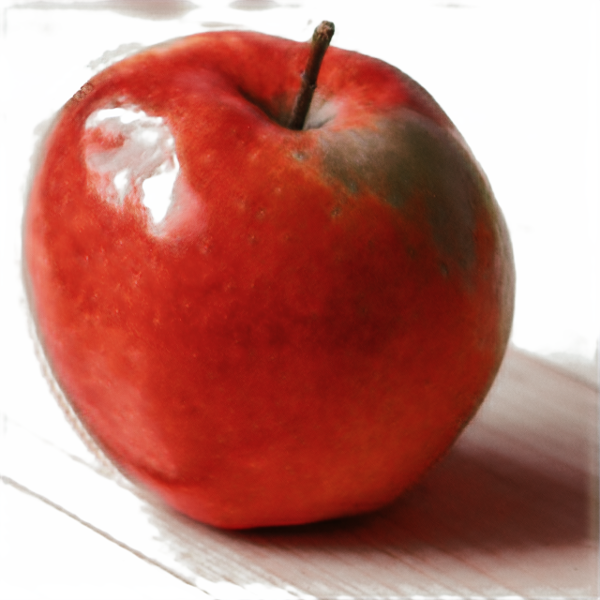}
		\captionof{figure}{low texture}
		\label{fig:lowtexture}
	\end{minipage}
	\begin{minipage}{.245\textwidth}
		\centering
		\includegraphics[width=.95\linewidth]{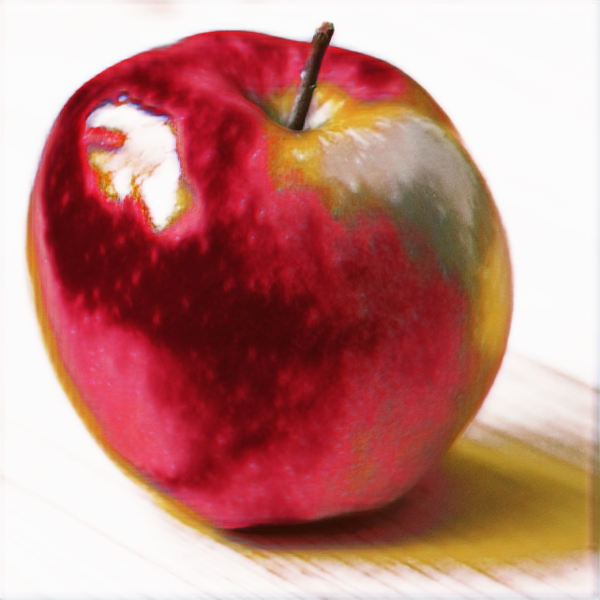}
		\captionof{figure}{high texture}
		\label{fig:hightexture}
	\end{minipage}
		\begin{minipage}{.245\textwidth}
		\centering
		\includegraphics[width=.95\linewidth]{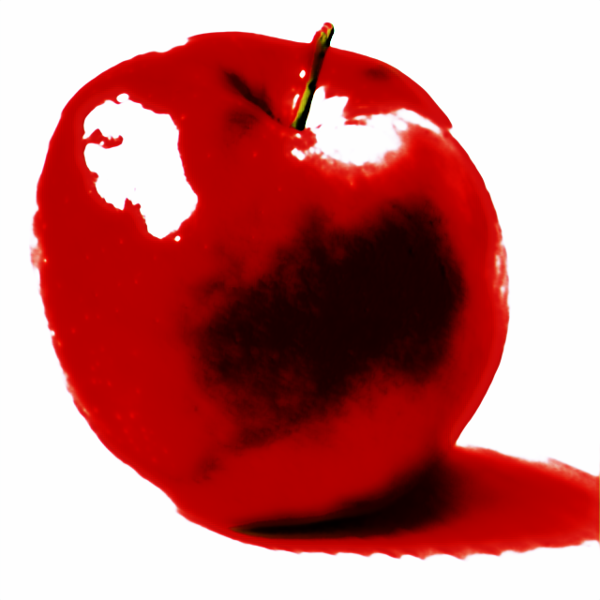}
		\captionof{figure}{low repetition}
		\label{fig:low repetition}
	\end{minipage}
	\begin{minipage}{.245\textwidth}
		\centering
		\includegraphics[width=.95\linewidth]{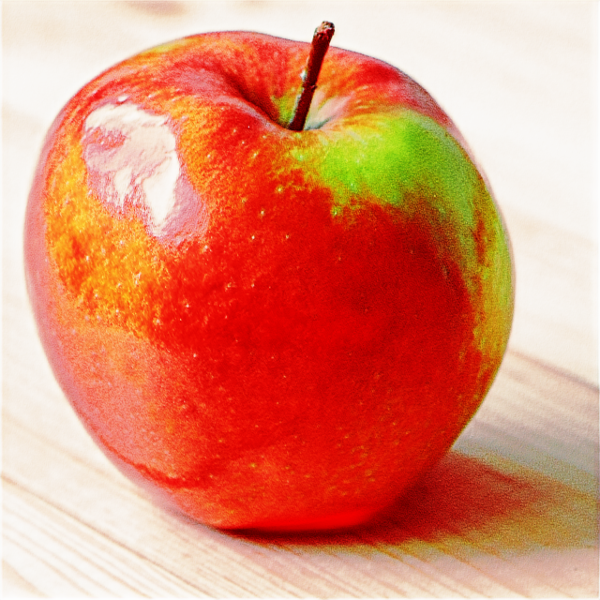}
		\captionof{figure}{high repetit.}
		\label{fig:highrepetition}
	\end{minipage}
\end{figure}

\begin{minipage}{0.245\textwidth}
    \captionsetup{type=figure} 
    \includegraphics[width=.95\linewidth]{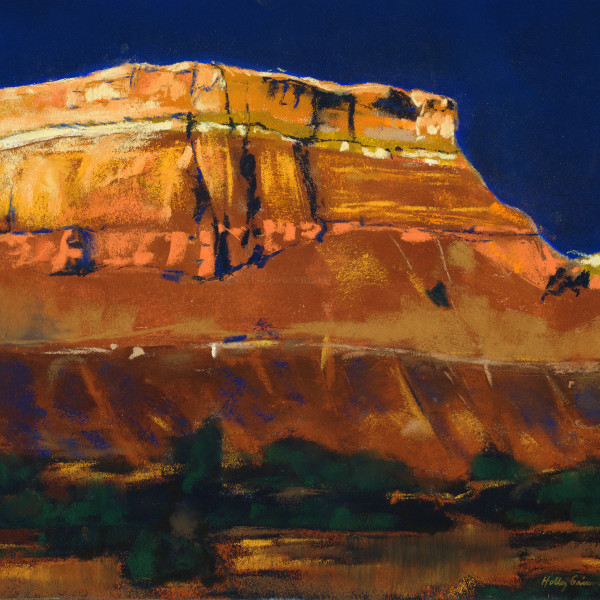}
    \captionof{figure}{original}
    \label{fig:art}
\end{minipage}
\begin{minipage}{0.245\textwidth}
    \captionsetup{type=figure} 
    \includegraphics[width=.95\linewidth]{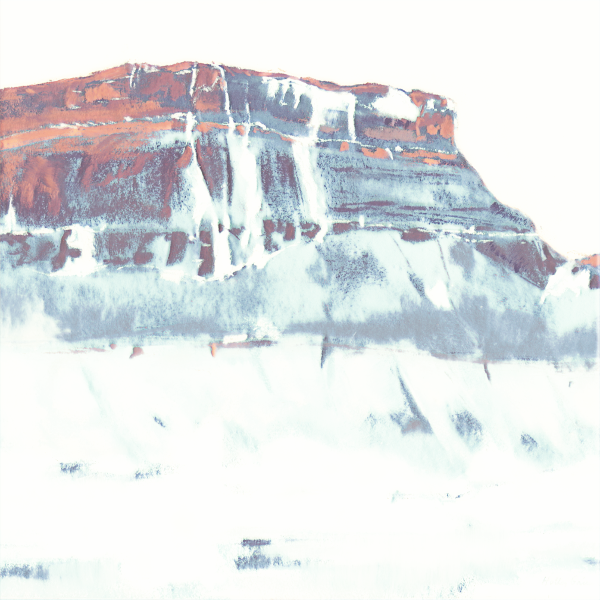}
    \captionof{figure}{translated}
    \label{fig:translated}
\end{minipage}
\begin{minipage}{0.245\textwidth}
    \captionsetup{type=figure} 
    \includegraphics[width=.95\linewidth]{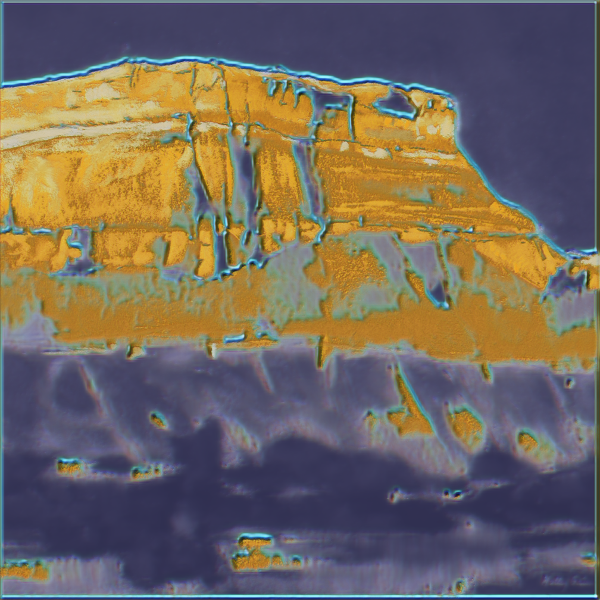}
    \captionof{figure}{reconstructed}
    \label{fig:reconstructed}
\end{minipage}
\begin{minipage}{0.245\textwidth}
The most surprising result of this project is the painterly effect that the ACAN was able to inject into the CycleGAN generated images as seen in Figures \ref{fig:translated} and \ref{fig:reconstructed}.
\end{minipage}

\subsubsection*{Acknowledgments}

This project was initially developed as part of the 2018 OpenAI Scholars program. I would like to thank my mentor, Christy Dennison from OpenAI, for her helpful comments along with support from Larissa Schiavo, Joshua Achiam, Jack Clark, and Greg Brockman from OpenAI.

\section*{References}

\small

[1] Zhu, J.\ \& Park, T.\ \& Isola, P.\ \& Efros, A.A.\ (2017) Unpaired Image-to-Image Translation using Cycle-Consistent Adversarial Networks. {\it arXiv:1703.10593}.

[2] Nichol, K. (2016) Kaggle dataset: Painter by numbers.\\ \url{https://www.kaggle.com/c/painter-by-numbers}.

[3] He, K.\ \& Zhang, X.\ \& Ren, S.\ \& Sun, J.\ (2015) Deep Residual Learning for Image Recognition {\it arXiv:1512.03385}.

[4] Malu, G.\ \& Bapi, R.S.\ \& Indurkhya, B.\ (2017) Learning Photography Aesthetics with Deep CNNs. {\it arXiv:1707.03981}.

\end{document}